%% file: main.tex
\documentclass[conference]{IEEEtran}
\IEEEoverridecommandlockouts

\usepackage[table,xcdraw]{xcolor}

\usepackage{booktabs}

\input{preamble.tex}

\def\redc{\cellcolor[HTML]{FF999A}}
\def\orangec{\cellcolor[HTML]{FFCC99}}
\def\yellowc{\cellcolor[HTML]{FFF8AD}}

\def\BibTeX{{\rm B\kern-.05em{\sc i\kern-.025em b}\kern-.08em
    T\kern-.1667em\lower.7ex\hbox{E}\kern-.125emX}}

\newcommand{\teaser}{
    \includegraphics[width=\linewidth]{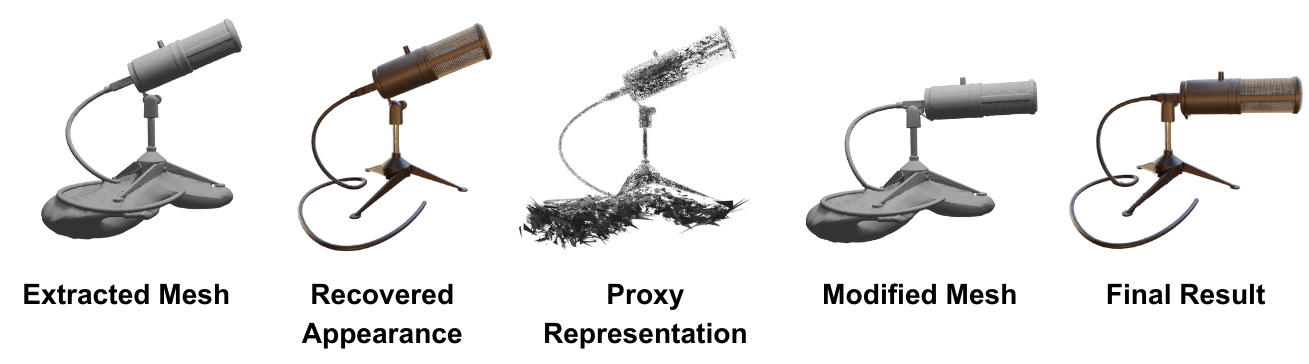}
    \vspace{0.5em}
    \par
    \parbox{\linewidth}{
        \footnotesize Fig 1. 
        We propose a novel method for editing 3D Gaussian Splatting (3DGS) scenes using an intermediate triangle soup proxy. 
        Our approach leverages a neural surface to guide 3DGS optimization, ensuring alignment between the appearance model and scene geometry. 
        This enables seamless propagation of manual mesh edits through the proxy, facilitating intuitive, mesh-guided appearance modifications. 
        Unlike previous methods relying on triangle soup proxies, our approach resolves connectivity limitations, expanding the range of possible edits and making the process more user-friendly.
    }
    \vspace{1em}
}

\makeatletter
\apptocmd{\@maketitle}{\centering\teaser}{}{}
\makeatother

\addtocounter{figure}{1}
    
\begin{document}

\title{Neural Surface Priors for Editable Gaussian Splatting\\
}

\author{
    \IEEEauthorblockN{Jakub Szymkowiak\textsuperscript{*1,2,\dag}, Weronika Jakubowska\textsuperscript{*3,\ddag}, Dawid Malarz\textsuperscript{1,4}, Weronika Smolak-Dy\.{z}ewska\textsuperscript{1,4},\\ Maciej Zięba\textsuperscript{3,5}, Wojtek Pałubicki\textsuperscript{2}, Przemyslaw Musialski\textsuperscript{1,6}, Przemysław Spurek\textsuperscript{1,4}}
    \\ 
    \IEEEauthorblockA{\textsuperscript{1}IDEAS NCBR, \textsuperscript{2}Adam Mickiewicz University, \textsuperscript{3}Wroc\l aw University of Science and Technology,\\
    \textsuperscript{4}Jagiellonian University, \textsuperscript{5}Tooploox, \textsuperscript{6}New Jersey Institute of Technology}
    \\ 
    \IEEEauthorblockA{\textsuperscript{\dag}jakub.szymkowiak@ideas-ncbr.pl, \textsuperscript{\ddag}weronika.jakubowska@pwr.edu.pl}
    \\ 
    \IEEEauthorblockA{\scriptsize \textsuperscript{*}Equal contribution}
}

\maketitle

\begin{abstract}
\input{sections/00_abstract}
\end{abstract}


\section{Introduction}
\input{sections/01_introduction}

\section{Related Work}
\input{sections/02_related_work}

\section{Method}
\input{sections/03_method}

\section{Experiments}
\input{sections/04_experiments}

\section{Conclusions}
\input{sections/05_conclusions}

\printbibliography

\vspace{12pt}
\color{red}


\end{document}

%% file: preamble.tex
\usepackage[T1]{fontenc} 

\usepackage{amsmath,amssymb,amsfonts}
\usepackage{algorithmic}
\usepackage{graphicx}
\usepackage{xcolor}

\usepackage[colorlinks=true, linkcolor=blue, urlcolor=cyan, citecolor=magenta]{hyperref}

\usepackage[backend=biber]{biblatex}
\usepackage{etoolbox}

\usepackage{afterpage}
\usepackage{mathtools}
\DeclareMathOperator{\diag}{diag}

\newcommand{\R}{\mathbf{R}}

\DeclarePairedDelimiter{\br}{(}{)}
\DeclarePairedDelimiter{\norm}{\|}{\|}
\DeclarePairedDelimiter{\abs}{|}{|}

%% file: sections/00_abstract.tex
In computer graphics and vision, recovering easily modifiable scene appearance from image data is crucial for applications such as content creation. 
We introduce a novel method that integrates 3D Gaussian Splatting with an implicit surface representation, enabling intuitive editing of recovered scenes through mesh manipulation.
Starting with a set of input images and camera poses, our approach reconstructs the scene surface using a neural signed distance field.
This neural surface acts as a geometric prior guiding the training of Gaussian Splatting components, ensuring their alignment with the scene geometry. 
To facilitate editing, we encode the visual and geometric information into a lightweight triangle soup proxy.
Edits applied to the mesh extracted from the neural surface propagate seamlessly through this intermediate structure to update the recovered appearance.
Unlike previous methods relying on the triangle soup proxy representation, our approach supports a wider range of modifications and fully leverages the mesh topology, enabling a more flexible and intuitive editing process.
The complete source code for this project can be accessed at \texttt{\href{https://github.com/WJakubowska/NeuralSurfacePriors}{github.com/WJakubowska/NeuralSurfacePriors}}.

%% file: sections/01_introduction.tex
Recovering scene appearance has been a longstanding challenge in computer graphics.
Neural Radiance Fields (NeRF) \cite{mildenhall2020nerf} and similar methods \cite{yu2021plenoxels, Muller2022ingp, chen2022tensorf} query neural networks, parametric grids, or a hybrid of those two, to sample color density along each ray, and employ volumetric rendering to produce high-quality appearance representations.
More recently, 3D Gaussian Splatting (3DGS) \cite{kerbl20233d} has gained significant attention for its fast training times and real-time viewing capabilities. 
3DGS captures scene appearance by optimizing each Gaussian's position in world space, scaling, rotation, and spherical harmonics coefficients, forming a mixture of thousands of local appearance estimators.

Despite its impressive results in novel-view synthesis tasks, 3DGS lacks the ability to recover accurate representations of scene geometry, producing only a noisy point cloud.
As a result, it is inherently incompatible with geometry sculpting tools commonly found in computer graphics software, which are primarily targeted towards editing connected meshes.
This limits the application of 3DGS in tasks such as scene modification, animation and physics simulations.

Several previous works have extended 3DGS, enabling surface extraction and scene editing.
One approach involves jointly optimizing the Gaussians and a neural signed distance field (SDF), and extracting a coarse mesh using the Marching Cubes \cite{marching-cubes, newman2006survey} algorithm. 
By directly binding the Gaussians to the mesh surface and further optimizing its vertices together with the appearance, methods such as \cite{guedon2024sugar} and \cite{gao2024manigs} allow for natural scene modifications, as manual mesh edits immediately alter the arrangement of Gaussians. 

Another line of work \cite{waczynska2024games} proposes to skip the mesh extraction step by encoding Gaussians in a triangle soup proxy representation, where each triangle corresponds to a single Gaussian.
Changes to the proxy directly affect the geometry of the Gaussians, resulting in modifications to the recovered appearance. 
This approach decouples Gaussians from the mesh surface, allowing greater flexibility in optimizing their positions and orientations.
However, it is less compatible with existing geometry modification pipelines since triangle soups lack the topological information present in meshes.
For example, connectivity of the mesh faces enables techniques like rigging, allowing for intuitive animation.
In contrast, triangle soups are primarily edited using grid-based methods such as lattice deform, which significantly limit the range and precision of possible modifications.

In this paper, we address this limitation by introducing a novel method to propagate mesh edits to the triangle soup proxy representation and, consequently, to the recovered scene appearance. Additionally, we propose a complementary pipeline that utilizes a neural signed distance field to recover the scene geometry, extract a mesh, and guide the training of 3D Gaussian Splatting.
This ensures that Gaussians are closely aligned with the scene geometry without being directly bound to its surface. 
With this setup, edits to the extracted mesh -- whether directly applied or after optional remeshing -- can be seamlessly transferred through the proxy to the recovered appearance while preserving visual consistency.
Furthermore, the high-fidelity mesh extracted by our pipeline enables the use of techniques that rely on detailed mesh geometries, such as physics simulations based on finite element methods.

In summary, our contributions are as follows:
\begin{itemize}
    \item A pipeline for geometry and appearance recovery from image data, designed to support triangle soup proxy editing workflows.
    \item A new method for scene editing, enabling mesh-guided transformations of the proxy to seamlessly propagate changes to the recovered appearance.
    \item An experimental evaluation showcasing the potential of our approach in novel view synthesis, geometry-based editing, and physical simulations. 
\end{itemize}

%% file: sections/02_related_work.tex
Recent advancements in computer graphics and vision have significantly enhanced our ability to recover and manipulate scene appearances. 
In this section, we provide a brief overview of the developments in the field, focusing on techniques that build upon 3D Gaussian Splatting for geometry recovery and scene modification.

\subsection{Novel View Synthesis}
Novel view synthesis is a task of generating images of the scene from unseen viewpoints based on a collection of ground truth views. 
Neural rendering methods, such as NeRF \cite{mildenhall2020nerf}, have leveraged deep learning methods to surpass the traditional Structure from Motion \cite{ozyesil2017survey-sfm, sfm-revisited} and Multi-View Stereo \cite{mvs-revisited} pipelines in terms of the visual fidelity of the reconstructed scenes. 
These techniques employ volumetric rendering, originally introduced in \cite{volume-rendering}, to sample densities along each ray originating from the viewpoint. 
While NeRF achieves impressive results, it requires processing numerous rays, resulting in slow training and costly inference. 
More recent works have alleviated these issues to a significant extent. 
Notably, InstantNGP \cite{Muller2022ingp} has introduced a differentiable hash-table encoding that allows for rapid training of the NeRF models.

In contrast to volumetric rendering-based methods, 3DGS \cite{kerbl20233d} models scene appearance using a point cloud of thousands of optimizable 3D Gaussian kernels, parametrized by location, orientation, and spherical harmonics-derived colors.
The kernels are then rasterized using a Gaussian rasterizer. 
This approach not only simplifies the rendering pipeline, but also significantly enhances training and inference efficiency, making it particularly suitable for real-time applications.


\subsection{Surface Reconstruction}
Numerous methods have been proposed to enhance both NeRF \cite{rosu2023permutosdf, yariv2023bakedsdf, wang2023neus, wang2023neus2} and 3DGS \cite{lyu20243dgsr, guedon2024sugar, chen2023neusg, dai2024surfels} to recover not only scene appearance but also its underlying geometry.
A common approach is to condition the opacity on the distance from the surface \cite{wang2023neus, lyu20243dgsr, wang2023neus2, chen2023neusg}, ensuring that high opacity values, which primarily contribute to rendering, align with the scene's geometry.
This often involves jointly optimizing a neural signed distance field (SDF) alongside the appearance model, followed by mesh extraction using Marching Cubes \cite{marching-cubes}. 

In particular, PermutoSDF \cite{rosu2023permutosdf} uses a dual architecture of two neural networks, where the color network is provided with the geometric information from the SDF network, to produce high-quality representations of the geometry and appearance. 
3DGSR \cite{lyu20243dgsr} extends 3DGS by conditioning the Gaussian kernels opacity on a jointly trained SDF with an additional volumetric rendering regularization. 

An alternative approach, used by SuGaR \cite{guedon2024sugar} and BakedSDF \cite{yariv2023bakedsdf}, strictly binds the appearance model to a coarse mesh obtained during an initial training stage, and optimizes it further to recover finer geometric details. 
Additionally, some methods replace 3D Gaussian kernels with 2D Gaussians \cite{dai2024surfels, guedon2024sugar, huang2dgs, waczynska2024games} positioned in planes tangent to the surface, offering a more natural geometric interpretation.

\begin{figure*}[t]
    \includegraphics[width=\linewidth]{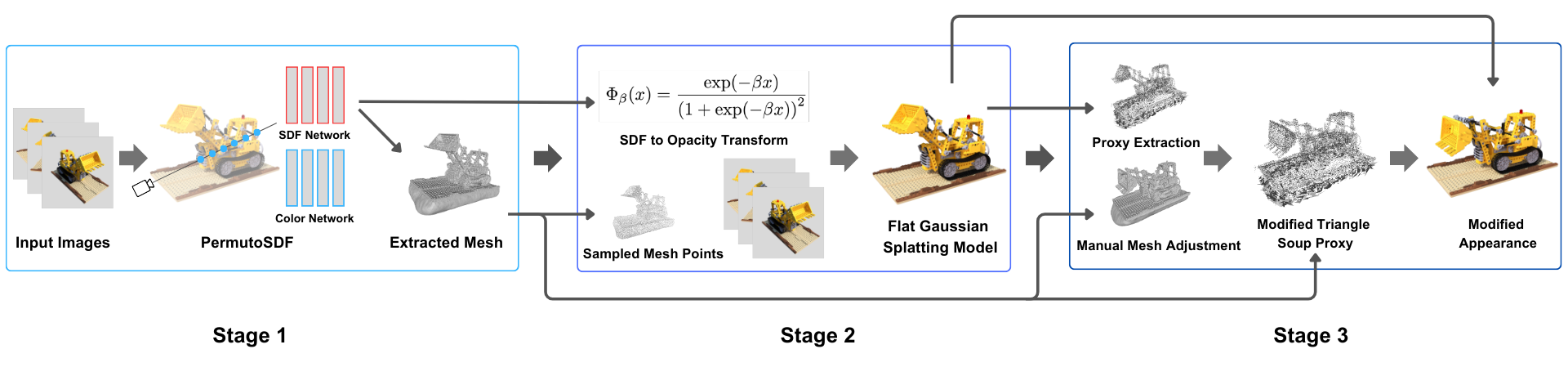}
    \caption{
        Schematic overview of our pipeline.
        (1) Starting with a collection of input images and corresponding camera poses, the initial stage utilizes PermutoSDF to generate a neural SDF and extract a mesh.
        (2) The second stage involves training a Gaussian Splatting model.
        The initial locations of kernels are sampled directly from mesh, and the opacity is defined by the minimum distance from the surface.
        (3) Given the modified mesh, the third stage involves the extraction of a proxy triangle soup representation, which allows for the propagation of the geometric changes onto this proxy. By recovering the updated Gaussian parameters from this modified proxy, a revised appearance representation is obtained.
    }
    \label{fig:pipeline}
\end{figure*}

\subsection{Scene Editing}
Editing and animating recovered appearance are crucial for content creation, interactive applications and capturing complex, dynamic phenomena. 
Given the abundance of highly-developed tools for editing, animating, and simulating physical interactions targeting mesh geometry representations, enhancing the recovered appearance with dynamical effects is inherently tied to the task of geometry extraction.
While numerous works have extended NeRF to support scene modifications \cite{park2021nerfies, pumarola2020dnerf}, 3DGS is more editing-friendly due to its point cloud-based representation of the scene.

Methods such as SuGaR \cite{guedon2024sugar}, Gaussian Frosting \cite{guedon2024gaussianfrosting} and Mani-GS \cite{gao2024manigs} bind Gaussians to a mesh extracted in the initial training stage.
As a result, any transformation applied to the mesh immediately alters the scene appearance, as the Gaussians move alongside mesh faces. 
However, the movement of Gaussians in the refinement stage is constrained by the mesh geometry.

A concurrent work \cite{huang2024gsdeformer} proposes a caged-based \cite{cages} deformation algorithm, which is suitable for any trained 3DGS scene and does not require additional optimization. 
Alternatively, \cite{huang2024scgs}  introduces control points for scene manipulation but requires additional video input.
PhysGaussian \cite{xie2024physgaussian} and GaussianSplashing \cite{feng2024gaussiansplashing} simulate physical phenomena, while \cite{yang2023deformable} employs deformable Gaussians to capture dynamic scenes from monocular videos.
As such, these methods do not offer general solutions for scene editing.

Finally, GaMeS \cite{waczynska2024games} and D-MiSo \cite{waczynska2024d} rely on a triangle soup proxy representation that encodes the recovered orientations and positions of the Gaussians into a set of disconnected triangles.
While conceptually simple, this approach limits the application of standard geometry-sculpting and animation tools such as rigging, making the editing process less intuitive. 
This specific limitation is the primary motivation for our work, which combines the simplicity of the triangle soup proxy structure with the versatility of mesh-guided editing to address issues caused by the lack of connectivity between the triangles in the proxy representation.

%% file: sections/03_method.tex
\begin{figure*}[t]
    \centering
    \includegraphics[width=0.85\linewidth]{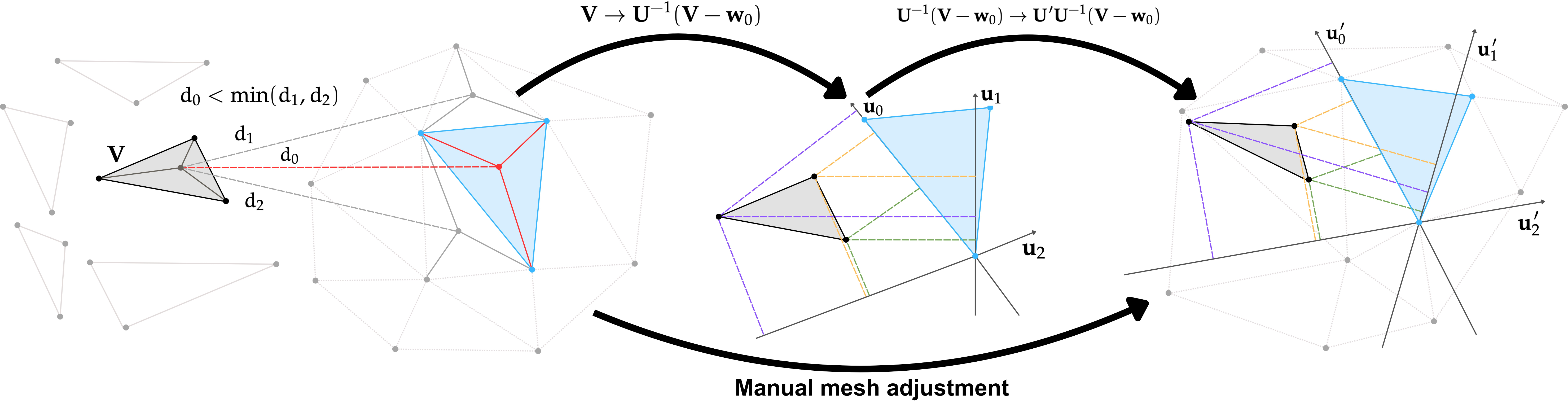}
    \caption{
        Visualization of propagating the mesh edit to a single proxy triangle. 
        We start by connecting the triangle $\mathbf{V}$ to the nearest mesh face (pictured in light blue).
        Then, its coordinates are expressed in the associated basis $\mathbf{U}$ (represented in the picture by gray axes pointing in the directions given by the basis). 
        A subsequent transformation aligns the triangle in the modified basis $\mathbf{U}'$, resulting in an updated representation of a single Gaussian’s shape and position.
    }
    \label{fig:modification-scheme}
\end{figure*}

Our mesh-guided appearance manipulation method requires a flat (or 2D) Gaussian scene representation similar to GaMeS \cite{waczynska2024games}, along with a mesh representing its underlying geometry. 
To incorporate scene modifications, we also assume a manually edited counterpart of the mesh with the same number of faces. 
To generate this data from input images, we propose a pipeline (illustrated in Figure \ref{fig:pipeline}) that first reconstructs the scene surface and trains a Gaussian Splatting representation guided by the reconstructed geometry. 
This is followed by the final editing stage that propagates mesh modifications to the recovered scene appearance through the triangle soup proxy representation. 
More specifically:
\begin{enumerate}
    \item First, we use PermutoSDF \cite{rosu2023permutosdf} to obtain a neural SDF model that guides the opacity of Gaussian kernels and enables mesh extraction with Marching Cubes for future editing.
    \item Second, we train a 3DGS-based model to recover the appearance from image data. In our method, we condition the opacity of the Gaussian kernels based on their distance from the neural surface and ensure each Gaussian lies flat on the surface by fixing one scaling parameter to a small value. Additionally, we incorporate a regularizer that aims to align Gaussian kernels' normals with the normals predicted by the SDF prior.
    \item Third, we encode the shapes and positions of each Gaussian into a triangle soup proxy. We associate each triangle in the proxy with the nearest face of the extracted mesh, and manually adjust the mesh by editing it in a geometry sculpting or animation software. Given the correspondence between the original mesh faces, and faces from the manually modified mesh, we are able to propagate the modification to the triangle soup proxy. Finally, we recover updated positions and orientations of the Gaussians, thereby altering the recovered appearance of the scene.
\end{enumerate}
The following sections explain each consecutive stage in more detail.

\subsection{Obtaining the Neural Surface Prior}
Our method starts with a collection of images representing a scene from multiple viewpoints and a set of associated camera poses. 
In the first stage, we aim to obtain a surface prior in the form of a neural network $f_{\theta} \colon \R^3 \to \R$ representing the signed distance to the object.
This is a crucial step as this neural surface guides the Gaussian Splatting optimization in the subsequent stage.
While other works propose to jointly train a neural SDF in addition to the appearance model \cite{lyu20243dgsr, chen2023neusg}, we opted to segregate it into a separate preprocessing stage, as it leads to a more detailed reconstruction of the geometry, and thus enhances the overall fidelity of later appearance modifications.

In particular, for this task we leverage PermutoSDF \cite{rosu2023permutosdf}. 
The PermutoSDF pipeline is comprised of two small multi-layer perceptrons learning to approximate the signed distance to the object and the view-dependent colors, respectively. 
Moreover, the color network uses as one of its inputs an additional geometric feature returned by the SDF network.
Subsequently, volumetric rendering is applied to generate the images.
The crucial strength of PermutoSDF lies in the multi-resolution tetrahedral lattice it employs to encode sampled rays.
This acceleration data structure is similar to the hash-tables used in InstantNGP \cite{Muller2022ingp}; however, unlike hyper-cubical grids, the number of lattice vertices scales linearly with dimensionality.
Consequently it improves training times, and provides the ability to efficiently represent spatio-temporal appearance and geometry.

\begin{figure}[t!]
\centering
    \includegraphics[width=0.8\linewidth]{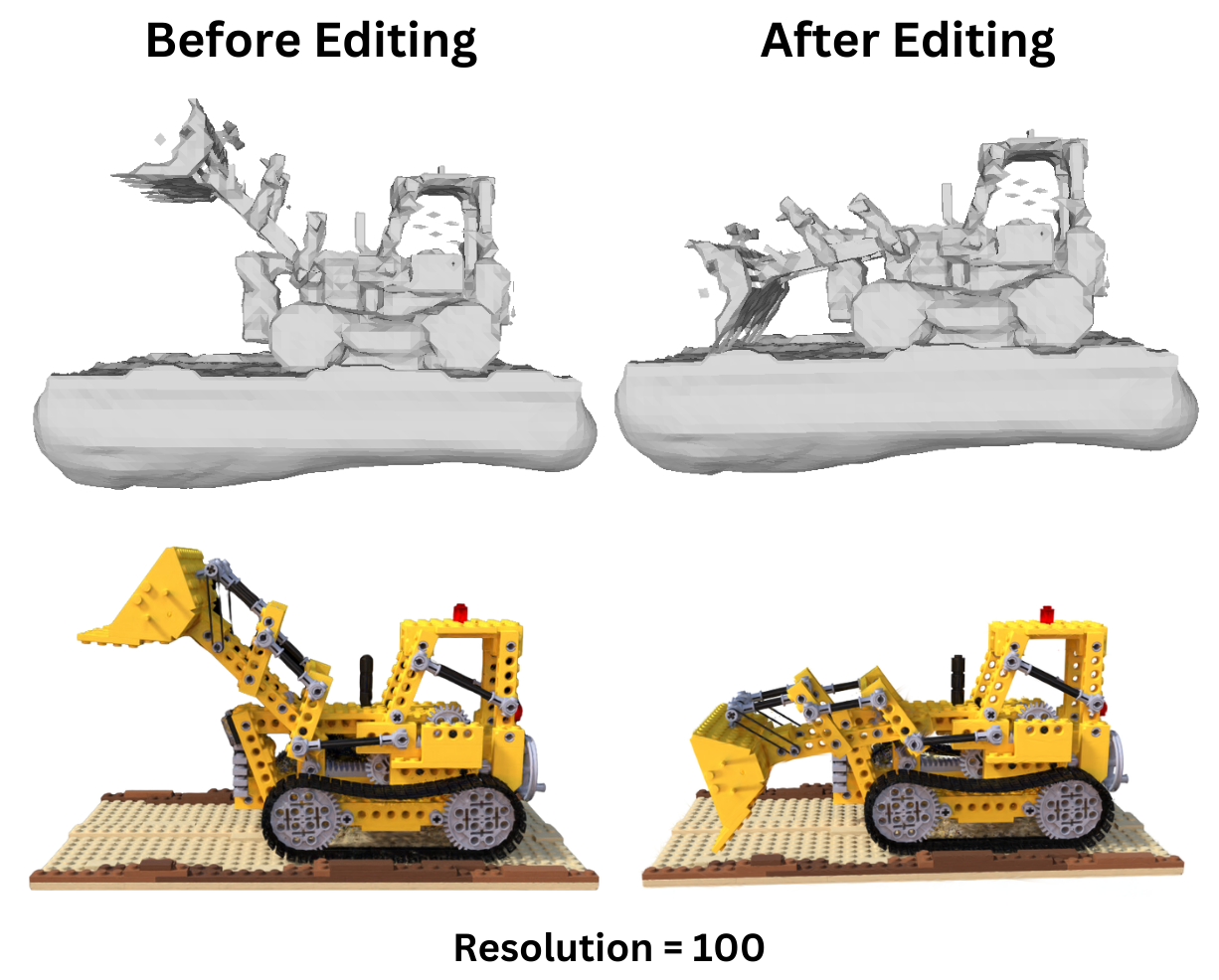}
    \caption{
        A modification created with our method on a low-resolution mesh.
        The left side shows the unmodified mesh and its render, while the right displays the modified version with the excavator's bucket lowered. 
        Despite the lower mesh resolution and applied edits, the output maintains high visual quality, comparable to the original high-resolution mesh.
    }
    \label{fig:edit_res}
\end{figure}

\subsection{Recovering the Appearance}
The goal of the second stage is to produce a Gaussian Splatting model of the recovered appearance suitable for mesh-guided editing.
3D Gaussian Splatting \cite{kerbl20233d} uses input image data and a point cloud obtained using SfM \cite{sfm-revisited} to produce an appearance representation comprised of many thousands of Gaussian kernels.
Each individual component is parametrized by a location vector $\mathbf{m} \in \R^3$, opacity value $\sigma \in [0,1]$, spherical harmonics colors, and a covariance matrix $\boldsymbol{\Sigma}$, which is decomposed as 
\begin{equation}
    \boldsymbol{\Sigma} = \mathbf{R} \mathbf{S} \mathbf{S}^{\top} \mathbf{R} \, ,
\end{equation}
where $\mathbf{S}$ is a diagonal scaling matrix, and $\mathbf{R}$ is a rotation matrix.
In practice, 3DGS optimizes a vector $\mathbf{s} \in \R^3$ such that $\mathbf{S} = \diag(\mathbf{s})$, and a quaternion $\mathbf{q}$ representing the rotation $\mathbf{R}$.
Each Gaussian component is then rendered using a Gaussian rasterizer to produce novel views of the scene.
Importantly, the number of Gaussians in the model adaptively changes during training, splitting or removing individual components based on their contribution to rendering the images.

For our purpose of mesh-guided scene editing, we aim to place each Gaussian component as aligned with the neural surface prior obtained in the first stage as possible. 
To this end, we modify the model described above in the following ways.
Start by denoting the neural network representing the recovered surface by $f_{\theta}$.
In our model, the opacity is not a learnable parameter, and instead is conditioned on the distance from the surface. 
More precisely, for each component, the opacity is a function of $f_{\theta}(\mathbf{x})$, where $\mathbf{x}$ denotes the location of the center.
Formally, $\sigma(\mathbf{x}) = (\Phi_{\beta} \circ f_{\theta})(\mathbf{x})$, where
\begin{equation}
    \Phi_{\beta}(x) = \frac{\exp(- \beta x)}{\br*{ 1 + \exp(- \beta x) }^2} \, ,
\end{equation}
with $\beta$ being a learnable parameter, is a bell-shaped function adopted from 3DGSR \cite{lyu20243dgsr}.
Intuitively, the higher the value of $\beta$, the better the alignment with the surface, although $\beta$ being too large leads to numerical instabilities.

Following GaMeS \cite{waczynska2024games}, we optimize two of the three scaling factors for each Gaussian, while fixing one to a negligible value of $\varepsilon = 10^{-8}$. 
This results in Gaussians being essentially two dimensional or flat.
Moreover, we incorporate the following regularizer to encourage the optimized normal direction of each Gaussian kernel to align with the normal predicted by the SDF:
\begin{equation}
    \mathcal{L}_{\rm{normal}}(\mathbf{x}) = \abs*{1 - \abs{\mathbf{n}(\mathbf{x})^{\top} \nabla f_{\theta}(\mathbf{x})}} \, .
\end{equation}
Here, $\mathbf{x}$ denotes the center of a single Gaussian, and $\mathbf{n}(\mathbf{x})$ is the corresponding normal, which in our case is the first column of the rotation matrix.
Finally, initial locations of the Gaussians are sampled from the mesh extracted in the first stage (cf. Fig \ref{fig:ablation}).
The rest of the pipeline remains the same as in 3DGS.

\subsection{Mesh-guided Appearance Modification}
The third stage propagates mesh modifications to the recovered scene appearance.
Given that the optimized Gaussians are flat, we can encode the information about their position and shape into a triangle soup proxy structure \cite{waczynska2024games}.

Suppose $\mathbf{m}, \mathbf{R}$, and $\mathbf{s}$ are, respectively, the location of the center, the rotation matrix, and the scaling vector of a single Gaussian.
A triangle $\mathbf{V}$ corresponding to this Gaussian is defined as an ordered set of its vertices $[\mathbf{v}_0, \mathbf{v}_1, \mathbf{v}_2]$, where $\mathbf{v}_0 = \mathbf{m}$, $\mathbf{v}_1 = s_1 \cdot \mathbf{r}_1$, and $\mathbf{v}_2 = s_2 \cdot \mathbf{r}_2$, with $\mathbf{r}_i$ being the $i$-th column of the rotation matrix $\mathbf{R}$.

Associating each Gaussian with a single triangle results in a triangle soup. 
Because this process is reversible, given a triangle soup, we can recover the shape and location of each Gaussian. 
More specifically, given a new face $\mathbf{V}'$, the recovered position $\mathbf{m}'$ is $\mathbf{v}'_0$ and the columns $\mathbf{r}'_0$, $\mathbf{r}'_1$, and $\mathbf{r}'_2$ of the recovered rotation matrix are given by:
\begin{equation}
    \mathbf{r}_0' = \frac{(\mathbf{v}_1' - \mathbf{v}_0') \times (\mathbf{v}_2' - \mathbf{v}_0')}{\norm{(\mathbf{v}_1' - \mathbf{v}_0') \times (\mathbf{v}_2' - \mathbf{v}_0')}} \, , \
    \mathbf{r}_1' = \frac{\mathbf{v}_1' - \mathbf{v}_0'}{\norm{\mathbf{v}_1' - \mathbf{v}_0'}} \, ,
\end{equation}
and $\mathbf{r}_2'$ is the orthonormal projection of $\mathbf{v}_2' - \mathbf{v}_0'$ onto the plane spanned by $\mathbf{r}_0'$ and $\mathbf{r}_1'$.
The corresponding scaling parameters are then $s_0 = \varepsilon$, $s_1 = \norm{\mathbf{v}_1' - \mathbf{v}_0'}$, and $s_2 = \langle \mathbf{v}_2' - \mathbf{v}_0' , \mathbf{r}_2' \rangle$.
We refer the reader to \cite{waczynska2024games} for more details.

To modify the appearance model optimized in the second stage, we first need to manually adjust the mesh extracted in the first stage, which we denote by $\mathcal{M}$.
This can be easily achieved using any of a vast array of geometry sculpting tools, such as Blender.
As a result, we obtain a modified mesh, denoted by $\mathcal{M}'$.
This modification can be then propagated to the triangle soup proxy as follows.

Observe that there is a one-to-one correspondence between the triangles composing $\mathcal{M}$ and the triangles composing $\mathcal{M}'$.
For each triangle $\mathbf{V}$ in the extracted triangle soup $\mathcal{T}$, we first find a triangle $\mathbf{W} \in \mathcal{M}$ such that the distance between the centers of the two is minimized.

Each face $\mathbf{W} = [\mathbf{w}_0, \mathbf{w}_1, \mathbf{w}_2]$ can be associated with an orthonormal basis $\mathbf{U}$, where the first two column vectors are given by
\begin{equation}
    \mathbf{u}_0 = \frac{\mathbf{w}_1 - \mathbf{w}_0}{\norm{ \mathbf{w}_1 - \mathbf{w}_0 }} \, , \
    \mathbf{u}_1 = \frac{(\mathbf{w}_1 - \mathbf{w}_0) \times (\mathbf{w}_2 - \mathbf{w}_0)}{\norm*{ (\mathbf{w}_1 - \mathbf{w}_0) \times (\mathbf{w}_2 - \mathbf{w}_0) }} \, ,
\end{equation}
and the third one, $\mathbf{u}_2$, is the cross product of those two. 
Similarly, we associate the corresponding face $\mathbf{W}' \in \mathcal{M}'$ with an orthonormal basis $\mathbf{U}'$.

In this way, we obtain a linear transformation $\mathbf{T} = \mathbf{U}' \mathbf{U}^{-1}$ that transforms $\mathbf{W}$ into $\mathbf{W}'$ (notice that the bases are orthonormal, and hence $\mathbf{U}$ is invertible).
Finally, we apply the associated transform $\mathbf{T}$ to each triangle $\mathbf{V} \in \mathcal{T}$, thus obtaining a modified triangle soup $\mathcal{T}'$.
Formally, each new triangle is given by
\begin{equation}
    \mathbf{V}' = \mathbf{T} \br*{ \mathbf{V} - \mathbf{w}_0} + \mathbf{w}'_0 \, ,
\end{equation}
where we first subtract the reference vertex $\mathbf{w}_0$ to recenter the triangle vertices, and then translate the modified triangle by $\mathbf{w}'_0$ to position it in its new location.
Note that the exact values of $\mathbf{T}$, $\mathbf{w}$ and $\mathbf{w'}$ depend on the triangle $\mathbf{V} \in \mathcal{T}$ being currently processed.

The updated triangles $\mathbf{V}' \in \mathcal{T}'$ are transformed into new locations and shapes of Gaussians using the procedure outlined above, while the rest of their parameters remain the same.
An illustration of the modification process for a single proxy triangle is provided in Figure \ref{fig:modification-scheme}.

%% file: sections/04_experiments.tex
\begin{figure}
    \centering
    \includegraphics[width=0.875\linewidth]{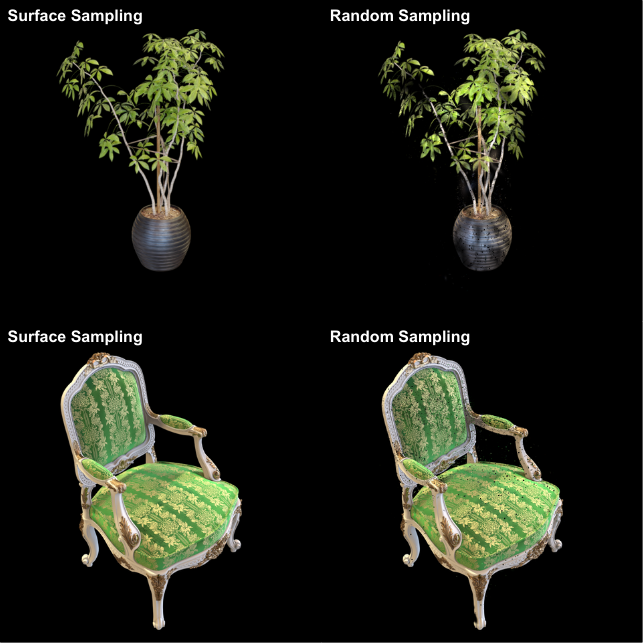}
    \caption{
        Comparison of rendering outputs for different Gaussian initialization schemes. Random initialization introduces artifacts, whereas sampling initial locations from the extracted mesh surface ensures cleaner and more accurate results.
    }
    \label{fig:ablation}
\end{figure}

\begin{figure*}[h!]
    \begin{center}
    \includegraphics[width=\linewidth]{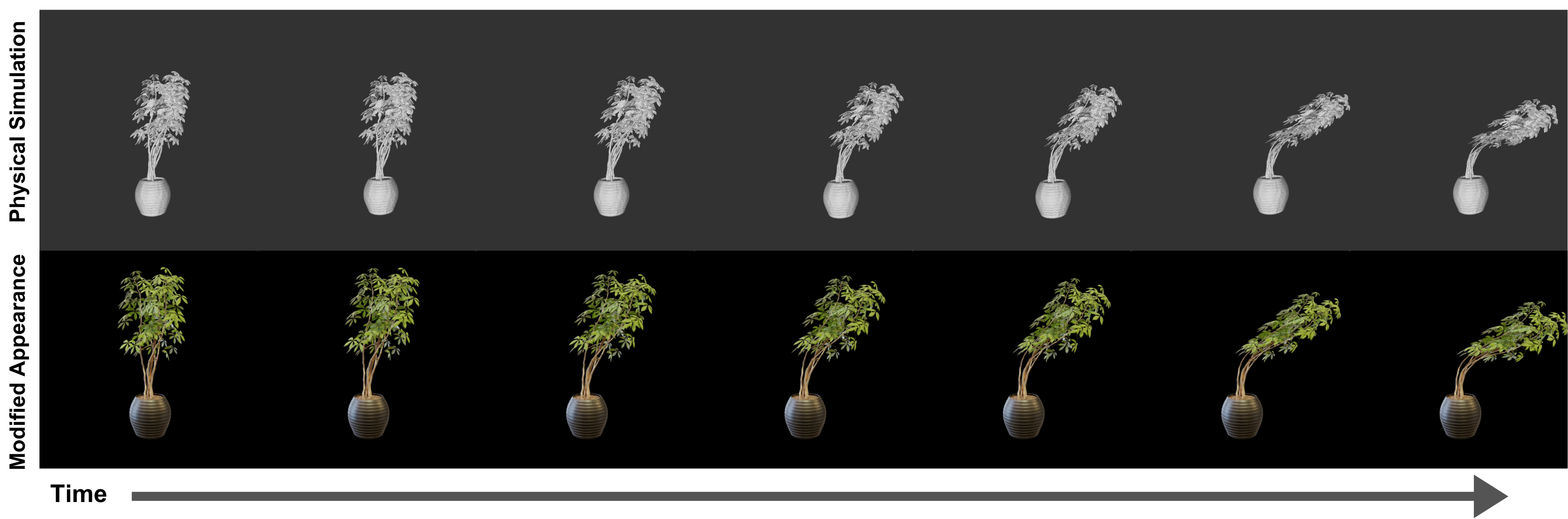}
    \caption{
        Progression of a wind-driven simulation of a Ficus plant. 
        Top row: changes in the object's geometry. Bottom row: corresponding modifications in appearance obtained using our method.
    }
    \label{fig:physics}
    \end{center}
    \vspace{-3mm}
\end{figure*}

\subsection{Numerical Results}
We assess the qualitative performance of our method in the tasks of geometry reconstruction and novel-view synthesis.

For novel-view synthesis, we evaluate the PSNR on the NeRF Synthetic dataset, which consists of eight manually designed scenes. 
Each scene includes $100$ training images with a resolution of $800 \times 800$ and associated camera poses.
Table \ref{tab:results_psnr_nerf} shows the results of this evaluation.
We compare our approach with NeRF \cite{mildenhall2020nerf}, Instant NGP (Ins-NGP) \cite{mueller2022instant}, Mip-NeRF \cite{barron2021mipnerf}, 3DGSR \cite{lyu20243dgsr}, NeuS \cite{wang2023neus}, NeRO \cite{liu2023nero}, BakedSDF \cite{yariv2023bakedsdf}, and NeRF2Mesh \cite{tang2022nerf2mesh}. 
Among SDF-based models, our approach outperforms others in the chair, drums, hotdog, lego, and mic scenes. 
Across all compared models, we achieve the highest PSNR for the mic and hotdog scenes.

Morever, we investigated the impact of mesh resolution on the recovered appearance. 
Table \ref{tab:results_res} presents the results of the PSNR, SSIM and LPIPS metrics across different mesh resolutions. 
The analysis indicates that metric values remain consistent regardless of the mesh resolution, demonstrating that render quality is unaffected by mesh granularity.

\begin{table}[t]
\centering
\scalebox{0.8}{ 
  \begin{tabular} {@{}l|*{8}{c}@{}}
    \toprule
    & Chair & Drums & Ficus & Hotdog & Lego & Materials & Mic & Ship \\
    \midrule
    NeRF     & 34.17 & 25.08 & 30.39 & 36.82 & 33.31 & \orangec 30.03 & 34.78 & 29.30 \\
    Ins-NGP  & \yellowc 35.00 & \orangec 26.02 & \orangec 33.51 & \yellowc 37.40 & \redc 36.39 & 29.78 & \orangec 36.22 & \redc 31.10 \\
    Mip-NeRF & \orangec 35.14 & \yellowc 25.48 & \yellowc 33.29 & \orangec 37.48 & \yellowc 35.70 & \redc 30.71 & \redc 36.51 & \yellowc 30.41 \\
    3D-GS    & 35.36 \redc & \redc 26.15 & 34.87 \redc & \redc37.72 & \orangec 35.78 & \yellowc 30.00 & \yellowc 35.36 & \orangec 30.80 \\
    \midrule

    NeuS & 31.22 & 24.85 & 27.38 & 36.04 & \yellowc34.06 & 29.59 \yellowc  & 31.56 & 26.94 \\
    NeRO & 28.74 & 24.88 & 28.38 & 32.13 & 25.66 & 24.85 & 28.64 & 26.55 \\
    BakedSDF & 31.65 & 20.71 & 26.33 &  36.38 \yellowc & 32.69 & \redc 30.48 & 31.52 & 27.55 \\
    NeRF2Mesh &  34.25 \yellowc & \yellowc 25.04 & \yellowc 30.08 & 35.70 &  34.90 \orangec & 26.26 & \yellowc 32.63 &  \yellowc 29.47 \\
    3DGSR & 34.85 \orangec &  \orangec 26.08 & 35.17 \redc  &  36.88 \orangec &  34.90 \orangec & \orangec 30.03 & \orangec 36.44 &  31.48 \redc \\
    Ours &  35.32 \redc &  \redc 26.09 & 34.32 \orangec & 37.77 \redc   & 35.42 \redc & 29.22 & \redc 36.62 & 30.64 \orangec \\
    \bottomrule
  \end{tabular}
  }
  \vspace{2mm}
  \caption{
    Novel view synthesis performance comparison.
    }
  \label{tab:results_psnr_nerf}
  \vspace{-5mm}
\end{table}

\begin{table}[t]
\centering
\scalebox{0.9}{
  \begin{tabular}{c|ccc|rrc}
    \toprule
    Resolution & SSIM & PSNR & LPIPS & Vertices & Faces & Time\,[s] \\
    \midrule
    200 &  0.9801 & 35.5185 & 0.01776 & 60361 & 120778  & 15.77\\
    400 &  0.9801 & 35.5331 & 0.01785 & 255931 & 512304 & 15.76\\
    600 &  0.9802 & 35.5607 & 0.01763 & 593934 & 1188712 & 19.82\\
    800 &  0.9802 & 35.5506 & 0.01765 & 1050086 & 2101122 & 22.93\\
    1000 & 0.9804 & 35.5663 & 0.01754 & 1667066 & 3334984 & 29.53\\
    \bottomrule
  \end{tabular}
  }
  \vspace{2mm}
  \caption{Comparison of metrics, vertex/face count, and edit time across mesh resolutions for the Lego scene.}
  \label{tab:results_res}
  \vspace{-5mm}
\end{table}

\begin{table}[t!]
\centering
\scalebox{0.8}{ 
  \begin{tabular} {@{}l|*{8}{c}@{}}
    \toprule
    & Chair & Drums & Ficus & Hotdog & Lego & Materials & Mic & Ship  \\
    \midrule
    Mesh &  35.32  &   26.09 & 34.32  & 37.77  & 35.42 & 29.22 & 36.62 & 30.64   \\
    Random & 28.07  &  24.71 & 27.53 &34.54  & 31.25 & 24.47 &	29.48 &  24.12\\
    \bottomrule
  \end{tabular}
  }
  \vspace{2mm}
  \caption{
    PSNR for different initialization schemes.
    }
  \label{tab:results_ablation}
  \vspace{-8mm}
\end{table}

Subsequently, we examined how mesh resolution impacts editing quality.
Figure \ref{fig:edit_res} compares an unmodified high mesh resolution scene with a modified appearance derived by editing a lower, $100$-resolution mesh.
Despite the reduced geometric detail, the appearance retains the visual quality; artifacts in lower-quality meshes do not affect rendering quality unless they directly impact the modifications.
Additionally, lower-resolution meshes reduce computational cost and speed up editing (cf. Table \ref{tab:results_res}). 
These findings demonstrate that our method enables efficient editing, regardless of mesh quality.



\subsection{Editing 3D Objects}
To illustrate the applicability of our approach, we conducted experiments on various scenes from the NeRF Syntetic and BlendedMVS datasets. BlendedMVS\cite{yao2020blendedmvs} consists of 17,000 training samples across 113 diverse scenes, including architectures, sculptures and small objects.

\begin{figure}[t]
    \centering
    \includegraphics[width=0.75\linewidth]{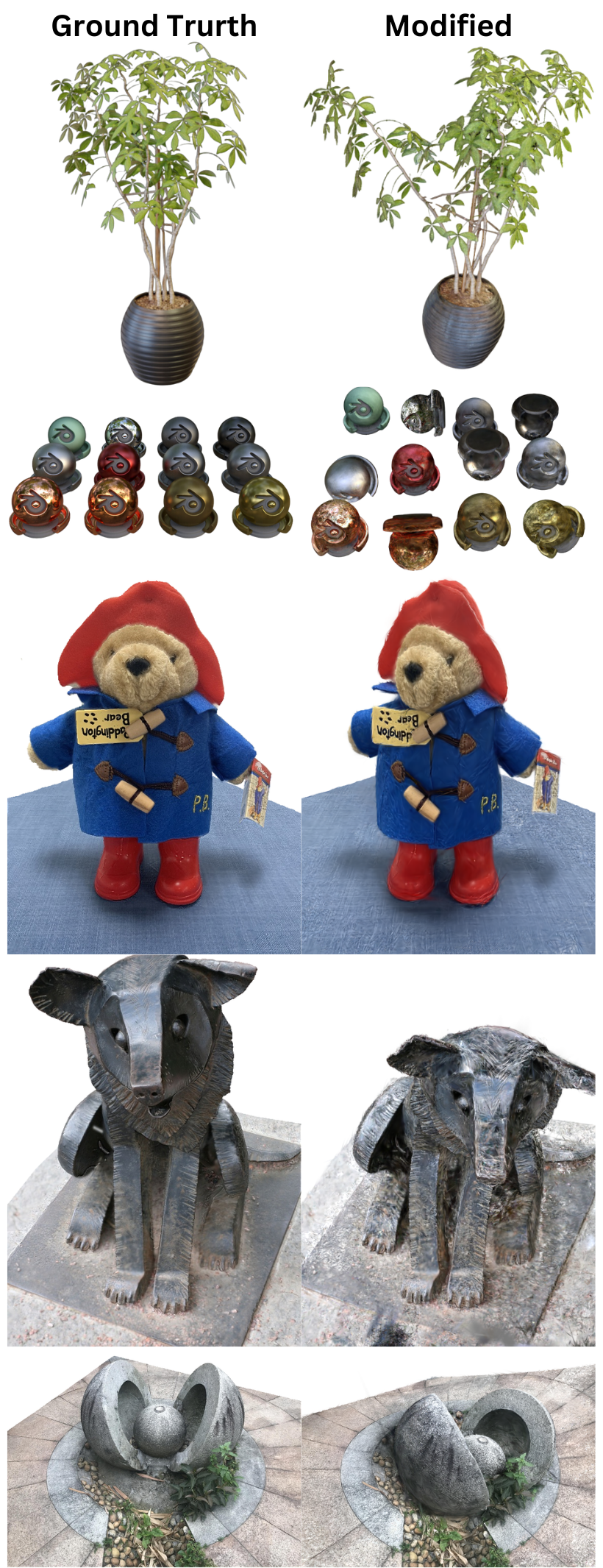}
    \caption{
        Example modifications produced with our method for scenes in the NeRF Synthetic and BlendedMVS datasets.
    }
    \label{fig:edits}
    \vspace{-5mm}
\end{figure}

We used Blender for mesh modification, and propagated the edits to the appearance using our method.
The results, shown in Figure \ref{fig:edits}, illustrate the application of various Blender tools (translate, rotate, knife, bevel, and randomize) tools applied to across different scenes. 

\subsection{Physics} 
We evaluate our method's performance in a dynamic scenario by simulating wind physics in Blender. 
This simulation models air movement affecting leaves and branches of the plant.
Using our method, we transferred the resulting dynamics from the surface of the mesh to the object’s appearance, as illustrated in Figure \ref{fig:physics}.
This demonstrates how our method consistently adapts the appearance of the represented object to dynamic deformations.

\subsection{Ablation Study}
Our method initializes Gaussians by sampling 100,000 points from the extracted mesh. 
To assess the impact of this initialization scheme, we instead initialized with 100,000 uniformly distributed points within a bounding cube.
Table \ref{tab:results_ablation} shows the evaluation results, showing that surface-based sampling yields higher PSNR scores.
Additionally, visual analysis of the rendered outputs (Fig. \ref{fig:ablation}) reveals that random initialization introduces noticeable artifacts.

%% file: sections/05_conclusions.tex



In this paper, we introduced a 3DGS-based scene editing method that enables mesh-guided modifications of the recovered appearance. 
Our approach ensures edits remain visually consistent by leveraging a scene representation trained in two steps: first, a a neural surface is trained to approximate scene geometry, which is then used as a prior for guiding the alignment of Gaussians.
The structure allows for flexible and intuitive editing while addressing the limitations of triangle soup representations, enabling a broader scope of topology-aware modifications.

While our method effectively propagates mesh edits to the recovered appearance, it is not without limitations. 
First, the underlying mesh must provide a structurally sound approximation of the scene geometry. 
First, the underlying mesh must provide a sufficiently consistent approximation of the scene geometry. 
While it does not need to be an exact match, issues such as disconnected components or floating artifacts in the geometry can introduce inconsistencies in the editing process, highlighting the importance of the neural surface prior.

Second, our approach assumes a fixed mesh topology. 
While transformations such as scaling, shearing, translation and rotation are well supported, editing operations that modify the number of mesh faces, such as remeshing or topology refinement, are not integrated into our framework, as we rely on maintaining correspondence between original and edited mesh faces.
Future work could explore ways to accommodate dynamic mesh topology.

Finally, certain edits may introduce appearance inconsistencies, particularly in lightning-dependent effects such as embedded shadows. 
For example, if an object is moved, its original shadow remains unchanged in the modified appearance. 
Addressing such inconsistencies would require relighting techniques, which are a natural extension of our approach.